\documentclass[letterpaper, 10 pt, conference]{ieeeconf}  

\IEEEoverridecommandlockouts                              
\usepackage{eucal}
\usepackage{amsfonts}
\usepackage{mathtools}
\usepackage{amsmath}
\usepackage{amssymb}
\usepackage{framed}
\usepackage{balance}
\usepackage{subfigure}
\usepackage{graphics} 
\usepackage{lipsum}
\usepackage{graphicx}
\usepackage{mathptmx} 
\usepackage{algorithm}
\usepackage{amssymb}  
\usepackage{algpseudocode}
\usepackage{booktabs}
\usepackage{tabularx}
\usepackage{booktabs}
\usepackage{array}
\usepackage{etex}
\usepackage{adjustbox}
\usepackage{adjustbox,lipsum}
\usepackage[table]{xcolor}
\usepackage{gensymb}
\usepackage{nicefrac}
\usepackage{mdwtab}
\usepackage{multirow}
\usepackage{colortbl}
\usepackage{hyperref}
\usepackage{color,soul}
\usepackage{epsfig} 
\usepackage{graphics} 
\usepackage{float}
\usepackage{bm}

\algrenewcommand\algorithmicforall{\textbf{foreach}}
\algrenewcommand\algorithmicindent{.8em}

\DeclareMathOperator*{\argmax}{arg\,max}  

\algnewcommand\algorithmicforeach{\textbf{for each}}
\algdef{S}[FOR]{ForEach}[1]{\algorithmicforeach\ #1\ \algorithmicdo}

\title{\LARGE \bf
KC-TSS: An Algorithm for 
Heterogeneous Robot Teams Performing Resilient Target Search
}

\author{Minkyu Kim$^{1,2}$, Ryan Gupta$^{1,3}$, and Luis Sentis$^{1,3}$
\thanks{The authors are with the $^{1}$Human Centered Robotics Lab, the $^{2}$Department of Mechanical Engineering, and the $^{3}$Department of Aerospace Engineering at the University of Texas at Austin}%
}

\begin{document}

\maketitle
\thispagestyle{empty}
\pagestyle{empty}


\begin{abstract} 

This paper proposes KC-TSS: K-Clustered-Traveling Salesman Based Search, a failure resilient path planning algorithm for heterogeneous robot teams performing target search in human environments. 
We separate the sample path generation problem into Heterogeneous Clustering and multiple Traveling Salesman Problems. 
This allows us to provide high quality candidate paths (i.e. minimal backtracking, overlap) to an Information-Theoretic utility function for each agent. 
First, we generate waypoint candidates from map knowledge and a target prediction model. 
All of these candidates are clustered according to the number of agents and their ability to cover space, or \emph{coverage competency}. 
Each agent solves a Traveling Salesman Problem (TSP) instance over their assigned cluster and then candidates are fed to a utility function for path selection. 
We perform extensive Gazebo simulations and preliminary deployment of real robots in indoor search and simulated rescue scenarios with static targets. We compare our proposed method against a state-of-the-art algorithm and show that ours is able to outperform it in mission time. Our method provides resilience in the event of single or multi teammate failure by recomputing global team plans online.
\end{abstract} 

\section{Introduction}

This study attempts to solve the problem of target search with multiple heterogeneous robots by generating informative global paths and executing search plans.
We interpret the target search problem as a variant of Coverage Path Planning (CPP) or exploration in which a team must additionally detect a missing or injured person or object in situations where mission speed is critical.
In traditional CPP a robot or a team of robots must find the optimal path such that the sensor footprint, or FoV, passes over the entire region, referred to as the search polygon. 
Exploration strategy is typically to find informative next view-point or path using sampling based approaches.
We take heterogeneous to mean different sizes of Field of View (FoV) and movement speeds.
The target may be considered as static and may be given with some prior target knowledge (e.g. a search and rescue mission searching near buildings while looking for people) 
and we assume static map information is known a priori.
When the target is static the proposed search algorithm simplifies to a CPP algorithm due to the uniform target prediction model over unseen space in the basic problem setting.

 \begin{figure}[t]
\label{fig:Sampling_PATHS}
\centering
\includegraphics[width=0.75\columnwidth]{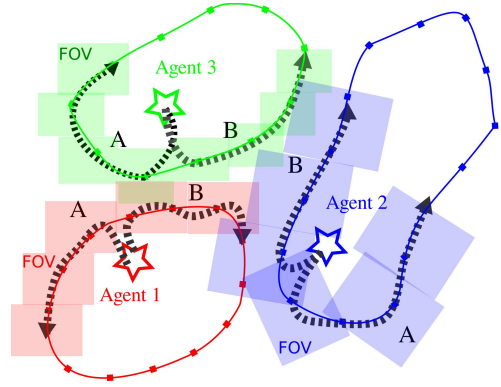}
\caption{Sampling paths along the TSP trajectory to calculate the Expected Information Gain. The three agents each have two candidate paths, A and B. Shaded rectangles denote the FoV of each agent at points along the path. If Agent 1 (red) selects path B then Agent 2 (blue) will choose path A to reduce overlap and increase $IG$.}
\end{figure}

CPP, exploration and target search are receiving significant attention from the robotics community due to the relevance in many important real-world scenarios including map building \cite{cao2021tare,caglioti2010,kumarplan2015,kumarmap2015}, cleaning/covering indoor areas \cite{tokekar2014placement,bahnemann2021revisiting}, security, surveillance and information gathering \cite{collins2021scalable,mccammon2021topological,hollinger2014sampling,cavinato2021dynamic}, reconnaissance or search and rescue \cite{pvenivcka2017reactive,kohlbrecher2014towards,hollinger2009efficient}. 



The goal of the presented algorithm is to generate high quality search plans for robot teams that attempt to utilize each team member to its full capacity. 
We define a metric, \emph{coverage competency}, which captures each member of the team's ability to effectively cover ground during search.
Our algorithm uses coverage competency to assign a search region to each agent through heterogeneous clustering (i.e. the spread of the waypoints reflects that agent's FoV size and movement speed). 
We also provide a new metric, Waypoint Allocation Factor (WAF), for measuring how evenly the waypoints were divided amongst agents based on their coverage competency score.

We demonstrate that our algorithm has proportional complexity in n and validate with n=50 agents. 
Further, we demonstrate that our algorithm is robust to failure of team members. 
If one or more agents fail, the remaining member(s) will complete the search. 
Because the server has access to robot plans over the entire search time via the TSP solution and current robot state, when one or more teammates fail, a re-planning step is invoked and nearby agents are assigned the failed members waypoints, as demonstrated in Fig. \ref{fig:recovery}. 


Overall, we present an algorithm for heterogeneous robot teams capable of handling dynamic and static targets. Our contributions are summarized as follows:
\begin{itemize}
  \item In the static target case, our method is equivalent to CPP (uniform target prediction) and we show it to outperform the state-of-the-art CPP method \cite{collins2021scalable} in mission completion time 
  \item We introduce a new parameter and metric for characterizing the success of heterogeneous teaming. Namely, \emph{coverage competency}, for quantifying agents' search skill and the metric, Waypoint Allocation Factor (WAF), for assessing the effectiveness of this parameter 
  \item Our algorithm can generate online mission plans for n robotic agents and monitors team progress to determine when re-planning is required.
  \item We validate our framework using both extensive Gazebo simulations of multiple robots and with a three robot team in a real-world environment
  \end{itemize}


This paper is organized as follows. Section II discusses related works. In section III we formally introduce our problem and describe our methods. In section IV we experimentally validate our methods in simulation finally section V concludes our work.

\section{Background and Related Work}
For decades CPP has received great attention for its relevance in navigation tasks. 
Early solutions offer an offline planner given a static map. A traditional method is to decompose the coverage region into cells using Boustrophedon decomposition \cite{choset2000coverage}. We employ a similar cell decomposition to generate the set of waypoints for complete coverage. 
A common strategy in this form of the problem is to break down the map and perform simple back-and-forth behaviors. 
One alternative to such behavior is to generate the set of points to visit and solve a Traveling Salesman Problem \cite{bahnemann2021revisiting, pvenivcka2017reactive}. 
\cite{milp2012} instead find the optimal order of visitation using mixed integer linear programming. 

There are also a number of geometric approaches \cite{tokekar2014placement, premkumar2020geometric}, which typically give global guarantees with respect to distance traveled during execution. However, such methods often make assumptions about perfect omni-directional sensing. 
Traditional methods for path planning are insufficient, however, when the goal is to operate in an online fashion to handle robot failure or dynamic environments.

Sampling based approaches are popular in mapping and exploration \cite{caglioti2010,kumarplan2015,kumarmap2015,hollinger2014sampling,dai2020}. Such methods account for sensor noise and environment complexity. Early work used frontiers \cite{yamauchi1997frontier} to explore spaces. Frontiers are the boundaries between explored and unexplored regions of the search polygon. The critical factor in information-theoretic based strategies is how to generate subsequent observation paths; that is, to provide quality candidates to the utility function. Early work \cite{caglioti2010} uses a laser range scanner to map a previously unknown small environment. 
In \cite{kumarplan2015,kumarmap2015} they leverage a utility function based on Cauchy-Schwartz quadratic mutual information to more efficiently generate plans to map 3D spaces. \cite{hollinger2014sampling} proposes three planning structures for information gathering missions like signal monitoring. 
\cite{dai2020} focuses on mission speed with MAVs by taking a hybrid approach to frontier and sampling based exploration strategies.

\cite{collins2021scalable} addresses the problem of scalable CPP by efficiently generating path plans with up to 150 agents in non-convex areas. 
They propose the robot team conducts auction and conflict resolution steps to determine the region of space they will cover.
We benchmark our method against this path planner, which is state-of-the-art in terms of coverage completion time \ref{table1}.
Our method features agent autonomy for search execution, a target prediction model, and automatic re-planning; however, when the target is static the path planning algorithm is equivalently a CPP problem, and therefore we use \cite{collins2021scalable} as a state-of-the-art comparison in terms of mission time and computation time (see Table \ref{table1}).

There are an abundance of real-time algorithms for multi-robot teams in coverage and exploration tasks. Groups have emphasized resilience to robot failure \cite{zhou2018resilient,rabban2020improved,ishat2021failure}, management of energy \cite{yu2019coverage,cai2021non,cesare2015multi} or communications \cite{woosley2020multi,shi2020communication,hollinger2012multirobot} constraints, and heterogeneous teaming \cite{sakamoto2020routing, kim2021information}. \cite{cai2021non} also considers an information gathering mission, however their future work lists increasing to n agent systems. \cite{cesare2015multi} performs search and exploration with UAVs, which are limited in both communication and battery life. They use a state machine to help team members decide between exploring, meeting, sacrificing and relaying. They leverage a frontier based method. In \cite{woosley2020multi} they consider a model of communication strength between the agents and a central control PC and cleverly use serial connection of the robots to maximize their exploration area.  
Instead of constraining the team to constant connectivity, \cite{hollinger2012multirobot} proposes a method of periodic communications at fixed intervals to update the full team. 
In \cite{hollinger2009efficient} an algorithm for solving the Multi-robot Efficient Search Path Planning (MESPP) problem for finding a non-adversarial moving target. This method provides theoretical bounds on search performance however it only scales up to five agents and does not provide resilience to robot failure. 

Our method, KC-TSS, builds on previous work \cite{kim2021information} where the team was not robust to agent failure because it did not provide complete search plans at any given moment. As a result of both this fact and the greedy planner, some of the generated paths were overlapping and lacked efficiency.
The additional clustering and TSP steps allow us to provide better candidate paths to the utility function and improve team cooperation. 

\begin{algorithm}[t]
 \caption{Multi-Agent Search()}  \label{waypoint}
 \renewcommand{\algorithmicrequire}{\textbf{Input:}}
 \renewcommand{\algorithmicensure}{\textbf{Output:}}
  \begin{algorithmic}
  \Require {$M, \mathbb{\mu}, SR$ \\ Entropy Map, Robot Pose, Coverage Competency}
  \Ensure {$P^*=\{p^*_1,p^*_2, \cdots, p^*_N\}$ (A set of paths)}
   \State $\mathcal{W} \gets ExtractionUnknownRegions()$  
   \State  $\mathbb{C}= HeterogenousClustering(\mathcal{W})$
    \For{$n \gets 1$ to $N$} \Comment  for each cluser in $\mathbb{C}$
   \State $r_n = TravelingSalesmanProblem(c_n)$
   \EndFor
    \For{$n \gets 1$ to $N$} \Comment  for each agent 
     \State $\hat{p_n} =SamplingPaths(r_n)$
     \For{$k \gets 1$ to $|\hat{p_n}|$}                    
     \State {$U(p_k) = IG(p_k)-c(p_k)$} \Comment Equation (7) 
     \EndFor 
     \State $p^{*} = p_k \gets \argmax{U(p_k)}$ \Comment{Get the best path} 
     \State $P^*.append(p^*)$
     \EndFor \\
  \Return $P^{*}$
 \end{algorithmic}
 \end{algorithm}

\section{Methods}


In this section we define a cooperative multi-agent target search algorithm for solving the problem setting described above. 
Given a search map (entropy map), a search polygon, the number of robots and initial robot position, this method first uses cell decomposition to generate a set of waypoints, which, if visited, provide complete map coverage over the unknown (high entropy) regions.
Then we perform a weighted clustering over all points and assign sub-regions (clusters) to each agent based on their coverage competency.
The next step solves instances of the single agent TSP in parallel to maintain the computational efficiency needed for generating global paths online if re-planning is necessary.
The result is two high quality candidate paths for each agent, that is, along both directions of the TSP solution (see Fig. \ref{fig:Sampling_PATHS}).
The final step is for these candidates for each agent to be considered by the IG based utility function (see Eq. \ref{eq:infogain_path}).
This IG is computed as entropy reduction along the path, where entropy is modeled by the target prediction model and unseen regions.
Because of the computational efficiency and global knowledge of this algorithm, re-planning is 
performed in event of agent failure or if some agents finish their assigned waypoints.


\subsection{Overall Framework}
\label{overall_framework}
Our search, presented in Algorithm \ref{waypoint}:
\begin{enumerate}
  \item Update search map using Bayesian filtering,
  \item Use Algorithm \ref{extract} to Generate waypoints from map,
  \item Heterogeneous Clustering using \emph{coverage competency}, See Algorithm \ref{HeterogeneousClustering},
  \item Solve Traveling Salesman Problem for each agent,
  \item Select optimal path using information-theoretic approach 
  \item If Re-plan conditions are met, repeat steps 2) - 5).
\end{enumerate}

\subsection{Target Estimation: Bayesian Filtering}
We use Bayesian Inference to recursively estimate target state $x$ through sequential observations $y$. Bayesian inference is commonly used to estimate target state in a probabilistic manner. This inference model aims to predict the posterior distribution of target position at time $k$, namely, $p(x_k)$. Bayesian filtering uses a prediction stage and a correction stage with incoming sensing information. Assuming that the prior distribution $p(x_{k-1})$ is available at time $k-1$, the prediction step attempts to estimate $P(x_k|y^{1:n}_{1:k-1})$ from previous observations as follows.
 
 \begin{equation}
    p(x_k|y^{1:n}_{1:k-1})=\int p(x_k|x_{k-1})p(x_{k-1}|y^{1:n}_{1:k-1})dx_{k-1},
\label{Eq:1}
\end{equation}
where $p(x_k|x_{k-1})$ is the target's motion model based on a first order Markov process. Then, when the measurement $y^{1:n}_k$ is available, the estimated state can be updated as 
\begin{equation}
    p(x_k|y^{1:n}_k)=\frac{p(y^{1:n}_k|x_k)p(x_k|y^{1:n}_{1:k-1})}{p(y^{1:n}_k|y^{1:n}_{1:k-1})}
\label{Eq:2}
\end{equation}
where $p(y^{1:n}_k|y^{1:n}_{1:k-1})=\int p(y^{1:n}_k|x_k)p(x_k|y^{1:n}_{k-1})dx_k$ and $p(y^{i:n}_k|x_k)$ is a sensing model for multi agent system, which can also be decomposed to each agent's sensing model $p(y^i|x)$. For the correction stage, the measurement of all agents are used to modify the prior estimate, leading to the target belief. If a static target is assumed the target prediction can be described as $p(x_k|x_{k-1})=\mathcal{N}(x_{k-1}; x_k, \Sigma)$, only containing a noise term with the previous target state. If it is a target motion model assumed to have some constant velocity, we can represent the model as $p(x_k|x_{k-1})= \mathcal{N}(x_{k-1}; x_k+V\Delta , \Sigma)$.
  
 \subsection{Target Prediction}
 The action plan our algorithm generates incorporates information from the target prediction model. The prediction model is made based on various prior knowledge or experience. For example, in a rescue mission, it might be useful to take advantage of the fact that people are more likely to be near collapsed buildings.In this paper, those are called context $c$ and can be used to estimate a target location. A Gaussian Mixture Model (GMM) can be constructed using a finite number of contexts at time $k$ using the following equation
  \begin{equation}
\begin{split} 
p(x_{k}|c_{k}) &= \sum^N_{i=1} p(x_{k}|c^i_{k})p(c^i_{k}) \\
               &= \sum^N_{i=1} \pi_{i}G(x|\mu_i,\Sigma_i)
\end{split}
\label{eq:xkck-1}
\end{equation}
 
 where $\pi_i$, $\mu_i$ and $\Sigma_i$ are a mixing coefficient, mean vector, and covariance for $i$-th distribution, respectively. We use a particle filter to implement this prediction model.

 \subsection{Search Map}
 The search map is updated using sensor data of each robot at each instance. The exploration begins under the assumption that the target exists in the search region. 
 We use a 2D occupancy grid map representation in which the search region is discretized into cells. More details of the search map cells are presented in \cite{kim2021information}. 
 A cell's occupancy status is used to measure the uncertainty of the target over the total search space (i.e the entropy map). Entropy is defined here as
 \begin{equation}
 H(M_t) = -\sum_{i=1}^{N}(m^i_t\log(m^i_t)+(1-m_t^i)log(1-m_t^i))
 \end{equation}
 where $m^i_t$ is the occupancy variable at time step $t$ and $N$ denotes the total number of cells. The search map is also maintained in this way. Furthermore, it can be filtered such that the degree of entropy increases over time in previously searched regions to account for a target prediction model.

\begin{algorithm}[t]
 \caption{ExtractionUnknownRegions()}  \label{extract}
 \renewcommand{\algorithmicrequire}{\textbf{Input:}}
 \renewcommand{\algorithmicensure}{\textbf{Output:}}
  \begin{algorithmic}
  \Require {$M, pos$ \Comment{(Entropy Map, Current Pose)}}
  \Ensure {$\mathcal{W}=\{w_1,w_2, \cdots, w_n\}$} NewUnknownSet
  \State $queue_m \gets \emptyset$
  \State $\mathcal{W} \gets \emptyset$
  \State $\text{Initialize} \; Visited[M_{m=0.5}]= False $
  \State $c_0 \approx M_{m=0.5}$ take the unknown cell 
  \State $Enqueue(queue_m, c_0)$
    \State $Visited[c_0] = True$
  \While{$queue_m$ is not empty()}{ \;
  \State $c_0 \gets DEQUEUE(queue_m)$ \;
  \For{$n_c \gets neighborhood(c_0)$}    \Comment{neighborcells}
    \If{$IsNewUnknown(n_c)$ and $visited[cell]==False$}
    \State {$w, Visited, l_c \gets buildNewUnknowns(n_c)$}
    \State {$\mathcal{W}.append(w)$}
    \State {$Enqueue(queue_m,l_c)$}
    \ElsIf{$Visited[n_c] == False $}
    \State $Enqueue(queue_m,n_c)$
    \State $Visited[n_c]=True$
    \EndIf
    \EndFor
  }
   \EndWhile \\
   \Return $\mathcal{W}(\text{NewUnknownSet})$
   \end{algorithmic}
 \end{algorithm}

 \subsection{Waypoint Generation} 
 Under the premise that we are given static map information a priori, the search map (entropy map) can be used to identify unexplored areas. We generate waypoints that divide these unexplored portions of the search polygon according to the unit FoV size (i.e. those points which have a 0.5 value). The resulting set of waypoints ensure complete coverage if visited. Our proposed algorithm generates this set using sensor range as shown in Algorithm \ref{extract}.

    \begin{algorithm}[t]
 \caption{HeterogeneousClustering()}  \label{HeterogeneousClustering}
 \renewcommand{\algorithmicrequire}{\textbf{Input:}}
 \renewcommand{\algorithmicensure}{\textbf{Output:}}
  \begin{algorithmic}
  \Require{$\mathcal{W}, \mathcal{\mu}_{1:n}, SR_{1:n}=\{\eta_1,\dots,\eta_n\}(\text{normalized})$ \\ (Waypoints, Centroids(Robot Poses), Sensing Capabilities)}
  \Ensure{ A Partition $\mathbb{C}=\{C_1,C_2, \cdots, C_n\}$}
    \State{$cost^{-}=0$}
    \State{$\text{Initiallize } \mathcal{\mu}$}
   \Repeat
      \State $\hat{C_i}=\{w_j: \eta_id(w_j, \mu_i) \le \eta_hd(w_j, \mu_h ) \; \text{for all } h=1,\dots,n \}$\\
      \Comment$ \text{assign all datapoints to the nearest cluster}$
       \State$\mu = \frac{1}{|C_i|}\sum_{w_j \in C_{i}}w_j$ \Comment$ \text{update centroids if required}$
       \State$ cost = \sum_{j=1}^n\sum_{w\in C_j}||\eta_j d(w,\mu_j)||$
      \State$\Delta cost = |cost-cost^{-}|$
      \If {$\Delta cost< \epsilon$}
      \State{$\mathbb{C} \gets \hat{C}$}
      \State {$break$}
      \EndIf
    \State {$cost^{-} \gets cost$} 
    \Until{$MAXLOOP$}\\
    \Return{ $\mathbb{C}$ }
 \end{algorithmic}
 \end{algorithm}

 \begin{figure*}
\centerline{\includegraphics[width=1.9\columnwidth]{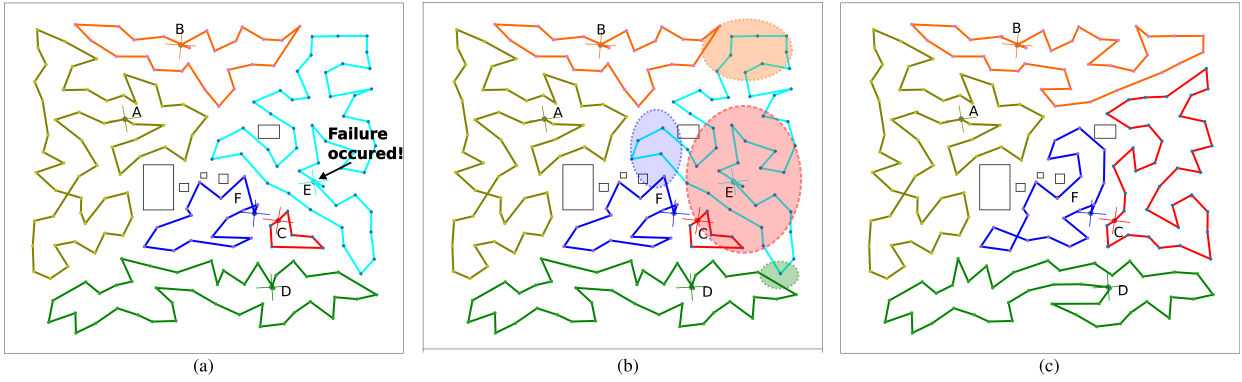}}
\caption{Scenario with 6 agents. All paths are described in different colors (a) Agent E (cyan) failure. (b) Re-planning step: Because the server has access to all robot plans, the waypoints assigned to the agent E can be reassigned to nearby agents. Ellipses represent the assignment of those waypoints (c) KC-TSS guarantees complete coverage by solving a new TSP instance.}
\label{fig:recovery}
\end{figure*}

 \subsection{Heterogeneous Clustering}
 We view the waypoint assignment problem in this case as a clustering problem. The K-means clustering method partitions a set of data into a predefined number of $K$ subgroups so as to minimize the sum of the distance squared between each data point and the centroid of assigned cluster. In the search setting $K$ is equal to the number of team members. Unlike in the conventional method, which is completely unsupervised learning to determine cluster centers, we perform clustering based on the known position of the robots. When computing the nearest neighbors based on robot location, we account for the robot's \emph{coverage competency}.
  
 Let $\mathcal{W}=\{w_1, w_2, \cdots, w_m\}$ where $w_i$ is $i$-th waypoint extracted from search map (\ref{extract}) and let $C=\{c_1, \cdots, c_k\}$ be a set of $K$ centroids, which correspond to the location of the agents. Suppose $K'$ ($K' \le K$) centroids are given as a constraint. The goal is to obtain the remaining number ($K-K'$) of centroids and to assign each point to the nearest centroid such that sum of the metric $d_j(x_1,x_2)=\eta_j|x_1-x_2|$ is minimized as follows.
  
  \begin{equation}
  \min_{c_{K'+1}, \cdots, c_K} \sum_{j=1}^K\sum_{w\in C_j}||\eta_jd_j(w,c_j)||
  \end{equation}
 where $\eta_j d_j(\cdot,\cdot)$ is the weighted distance, which reflects the $j$-th robot's sensing capability using the normalized coefficient $\eta_j$. We define \emph{coverage competency} $\eta_j$ as $\frac{SR_{min}}{SR_j}$ where $SR_j$ is the sensing capacity of $j$-th agent, which is the product of a moving speed and sensing range of agent $j$. Because we normalize this coefficient with $SR_{min}$, the minimum value of sensing capacity among all agents, we can use it to consider relative proximity between centroids and points. In this way we ensure that robots which are fast or have a large FoV are allocated more waypoints than robots which are not.  
  
  Furthermore, this formulation has the advantage of being flexible in varying situations. When the robots are evenly distributed in the search region, computed waypoints will easily be distributed among the agents. However, some situations arise in which robots start next to one another. If two agents with competency gaps begin next to each other, then all of the nearby points will be assigned to the more competent teammate. To avoid this, we compute a new centroid when any agent is assigned fewer than some minimum defined number of points. If, on the other hand, two agents of equal skill begin next to each other we compute the distance to other agents and determine which member is given the new centroid and in addition those nearby agents are given a new centroid. As a result of clustering, we receive the waypoints assigned to each agent (centroid) in $\mathbb{C}$.

  \subsection{Traveling Salesman Problem}
 After constructing and assigning a cluster to each robot we solve an instance of the Traveling Salesman Problem (TSP) separately for each agent in parallel. 
 In this study we adopted the Genetic Algorithm for solving TSP \cite{sedighpour2012effective}. In this setting the start point is the cluster's centroid (i.e. the robot's starting position). The result is the optimal route along the cluster. While this generates a set of $k$ optimal paths when considered independently, the solution does not consider the other agents. To overcome this lack of consideration of the team's combined effort, we generate a set of sample paths for each agent then use the information-theoretic framework using the aforementioned constraints. We exploit the utility function which maximizes the acquisition of information for the full team in a limited time while penalizing traveling costs. In this way we have provided higher quality samples to the utility function when compared to traditional or frontier-based methods.


   \begin{figure*}[t]
\centerline{\includegraphics[width=1.91\columnwidth]{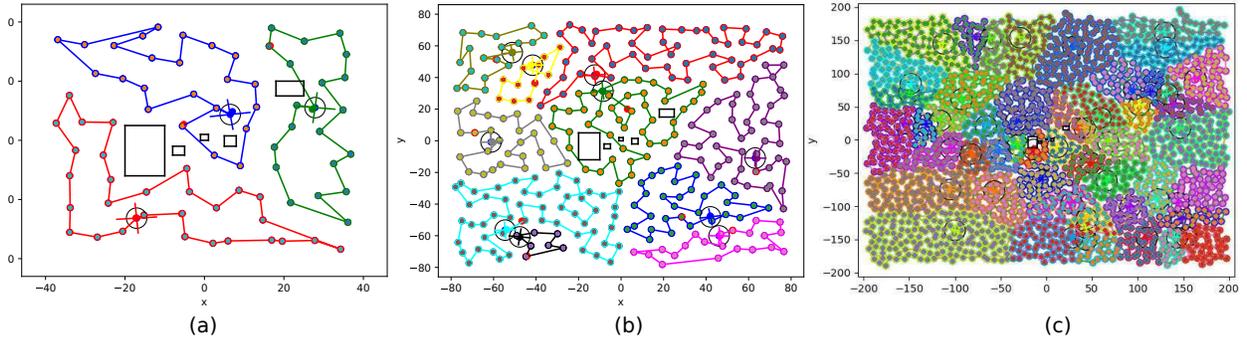}}
\caption{Simulation results with different number of agents in different size search regions. (a) 3 agent case (40$\times$40) (b) 10 agents (80$\times$80) (c) 50 agents (200$\times$200)}
\label{fig:tworobot_search}
\end{figure*}

  \subsection{Path Selection} 
   In this stage, our framework determines which direction the robots will travel along the given route from TSP (See Fig. \ref{fig:Sampling_PATHS}. Given the routes from TSP, candidate paths are sampled using several points which are close to robot positions. An A* planner is used to generate obstacle-free paths and all paths are re-parameterized with respect to agent's moving speed. Based on the number of sampled points along a path, we compute the Information Gain, denoted $IG(s)$, with the following equation:
   
 \begin{equation}
\begin{split}
IG(s) \approx  \sum_{i=1}^{N_{\text{s}}}H(FOV(\text{s}^i)) \qquad \qquad \qquad \qquad \qquad \qquad \qquad \qquad\\
=\sum_{i=1}^{N_{\text{s}}}\left[\sum_{j=1}^{N_c}\big[ m_{i,j}log(m_{i,j})
+(1-m_{i,j})log(1-m_{i,j})\big] \right]  \qquad  \quad \\
 \end{split}
 \label{eq:infogain_path}
\end{equation}
where $\text{s}^i$ denotes the $i$-th sampled point and $m$ is the occupancy probability, while $N_{\text{s}}$ and $N_c$ are the number of sampling points and the number of cells in the FOV given the sampled points, respectively. Thus, The $IG(\text{s})$ is calculated by summing over the FOV regions defined by sampled points through the path. The overall expected utility, $\mathbb{E}$[$\tilde{U}$], is then computed for the full team as

\begin{equation}
\mathbb{E}[\tilde{U}(x_t|y^1,y^2,\cdots,y^n)] = \sum_{i}^{n}\left( IG(\text{s}_i)-c(\text{s}_i)\right)
\end{equation}

where $c(s_i)$ denotes the traveling cost to move along the path $s_i$.

 \subsection{Re-planning} 
 In the event of robot failure or if some of the agents finish covering their sub-region without finding the target, it is desirable to re-plan for the team. When a preset percentage of the team covers their region we generate a new set of waypoints over the full unexplored search map and perform the full process again. Similarly, if a robot loses communication with the server or fails to move for an extended period of time, we perform the same process of resetting to generating waypoints ( step (2) in section \ref{overall_framework}). We explicitly define re-planning conditions. 
 \begin{itemize}
   \item Failure (stop operating or lost communication)
   \item Another prediction prediction model applied 
   \item All way points assigned are visited by the corresponding agent 
 \end{itemize}
 
 These re-planning conditions allow our team to be adaptable to changes in the environment including robot failure. One re-planning scenario is shown in Fig. \ref{fig:recovery}.

\subsection{WAF}
 We introduce a new metric called $WAF$, \emph{Waypoint Allocation Factor} to evaluate the contribution of the task according to the coverage competency. In order to compute how evenly the area coverage is allocated, we take the total swept area of each agent and divide by the \emph{coverage competency}. Specifically, we apply the following equation,
 \begin{equation}
 WAF = std\left(\lambda\frac{A(|r_i|)}{\eta_i}\right)
 \end{equation}
where $A(|r_i|)$ denotes total swept area for agent $i$, $\eta_i$ is the coverage competency, and  $\lambda$ means a normalizing constant. A value close to zero indicates that the waypoints were evenly distributed among the team after competency considerations

\subsection{Computational Complexity}
 Achieving computational efficiency is critical for achieving online planning as the number of agents increase. We analyze here the computational efficiency of the proposed algorithm using the following parameters; Number of cells $N_g$ (map size $/$ resolution), the number of trajectories to sample $N_t$, the number of agents $k$, and other parameters for other algorithm. 
 
 In detail, the complexity of the waypoint generation algorithm \ref{extract} is $O(N_g)$. The clustering has a time complexity of $O(ikmd)$ where $i$ denotes the fixed number of iterations (max iteration), $k$ is the number of clusters which is equal to the number of agents,  number of $m$ data and $d$ dimension of the data. It can be computed as  $\frac{N_g}{kN_{FOV}}$  and represents the worst case of decomposition (i.e. that the search region is totally unknown and divided by the product of number of agents and unit cell size of the FOV). The solution of the TSP with the genetic algorithm is of order $O(jn_0n^2)$ where $j$ is number of outer iterations of genetic algorithm and $n_0$ is the initial size of population, and $n$ is number of locations. Sampling based path selection algorithm takes $O(N_t \log{N_t}$ complexity. Overall, the complexity of our algorithm is $O(N_g +2ik\frac{N_g}{kN_{FOV}} +kjn_0n^2N_t\log{N_t})$.

\section{Experiments and Results}
\subsection{Numerical Simulation Results}
 We present python-based simulation results of our proposed approach to demonstrate its scalability. Agent state is represented as $s=[x,y,\theta]$ and has two control inputs $u=[v,\omega]$ as per the equations of motion for a non-holonomic mobile agent. Each agent is equipped with a ray sensor which has square-type of FoV with limited range. It is assumed that the simulation environment (search region) and all the static obstacles have a rectangular shape and obstacles can not be known in advance. To achieve robust collision avoidance, we use dynamic window approach \cite{fox1997dynamic} to generate each agent's control input. We tested different initial conditions with varying number of agents and we show some example scenarios in Fig \ref{fig:tworobot_search}.

 \begin{figure}[t]
\centerline{\includegraphics[width=0.95\columnwidth]{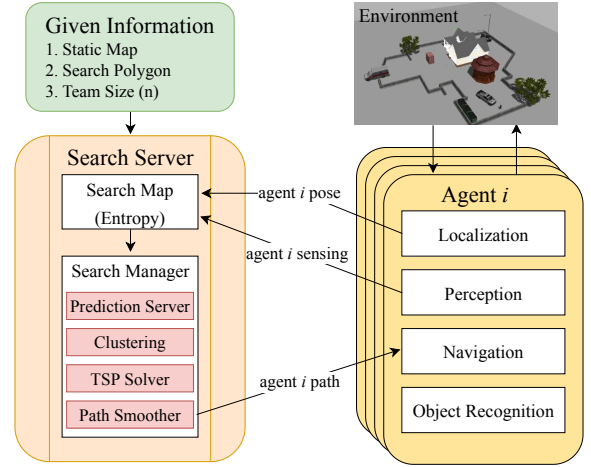}}
\caption{Flowchart describing simulation implementation and interactions between the search server and agents.}
\label{fig:api}
\end{figure}

\begin{figure*}[t]
\centerline{\includegraphics[width=1.8\columnwidth]{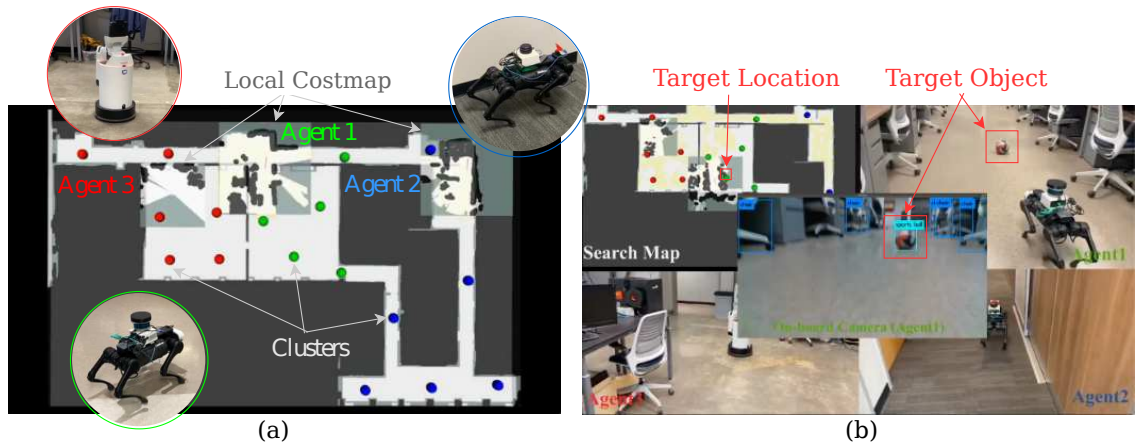}}
\caption{(a) Experimental validation for a three robot search in the Aerospace Building 4th floor. In the top left figure, the yellow regions are those which have been explored, while the white areas are regions of uncertainty. The target location is described as a red box and is unknown to the robots. The red, green and blue markers represent the clusters for each agent. (b) The completion of the experimental validation with Agent 1 converging to the target of interest. The object recognition is performed using the well-known YOLO algorithm to detect people (shown in the cyan box).}
\label{fig:asesearch}
\end{figure*}

\subsection{Comparative Results}
We benchmark our method against a state-of-the-art algorithm with the results shown in Table I. KC-TSS is shown to outperform the offline algorithm SCoPP \cite{collins2021scalable} when using the same search setting. That is the simulation trials were performed in the python-based numerical simulation in a region of the same size using the same number of agents which are given equal skill. We generate this data using 13 agents in the same size rectangular search polygon as presented by \cite{collins2021scalable} and we generate this data from 10 comparative trials.
 
  \begin{table}[t]
\caption{Comparative test}
\centering
\begin{tabular}{c||c}
\hline
\hline
Method & Completion Time (s) \\
\hline
Guastella's  & 3591.8 $\pm$ 0.00   \\
SCoPP  & 196.53 $\pm$ 14.53   \\
KC-TSS (ours) & 180.22 $\pm$ 21.23 \\
 
\hline
\hline
\end{tabular}
\label{table1}
\end{table}

 \subsection{Effectiveness of Coverage Competency} 
 
\begin{figure}[t]
\centerline{\includegraphics[width=0.7\columnwidth]{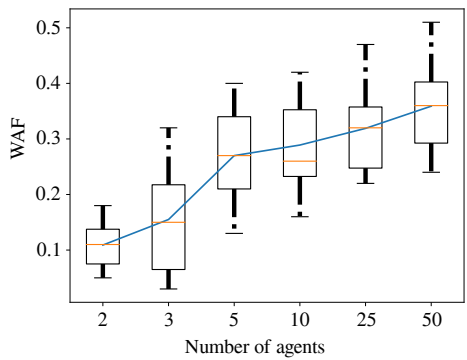}}
\caption{Waypoint Allocation Factor versus number of agents}
\label{fig:task allocation}
\end{figure}

 In simulation we tested six cases. Because of the random distribution of robot starting place, it will not approach zero because we cannot perfectly account for coverage skill. However, our results suggest that this value is safely maintained below 0.4 in all scenarios. It is natural that $WAF$ increases with the number of agents due to the random initial conditions. The data from these trials is in Fig. \ref{fig:task allocation}.
 
\subsection{Resilience to Failure}
One significant advantage of our framework is that it can quickly re-plan navigation behavior when one or more team members fail via loss of communication or navigational error (e.g. a computer loses internet connection, a legged robots falls). When faced with one of these events, we regenerate the set of waypoints over the unexplored region, cluster and allocate them, and then continue the method as before. Fig. \ref{fig:recovery} demonstrates this resilience to failure. 

\subsection{Gazebo-ROS Simulation}
We validate our framework's ability to transfer to robotic systems in human environments with high fidelity Gazebo simulations. The first is a 10mx20m simulation of the Anna Hiss Gymnasium apartment, the second is a 40mx50m region simulating the outside of a home environment (see Fig. \ref{fig:api} upper right corner) and the final is a 100mx100m town that has been struck by natural disaster. We validate the static target search capability in all three environments. In the town map, the agents search for an injured person and we use this information to guide the target prediction model by assuming the injured person will be near a building as opposed to in open space. We also demonstrate the resilience of the search method by performing re-planning under agent failure. This paper's accompanying video features the described scenarios.

To validate the inclusion of coverage competency, we present \ref{table2} to compare mission time for the agents when coverage competency is used versus when it is not used. Additionally, for the trials using coverage competency we include the $WAF$ score. This table uses teams of $n$ heterogeneous robots and is computed over 10 trials on each map with varying initial conditions. In all cases, it was confirmed that the search time was reduced and the WAF value was also low. There is a trend that the WAF increases slightly as the size of the map increases. The results clearly demonstrates that the addition of coverage competency impacts the search speed of the team and that the search area was more properly distributed.

 \begin{table}[b]
\caption{ Search Time (s) with and without Coverage Competency}
\centering
\begin{tabular}{c||c|c ||c|c}
\hline
\hline
Map (\# robots) & w/o CC & WAF  & w CC (proposed) &WAF \\
\hline
Apartment (2)  & 124.8 $\pm$ 13.2 & 0.23 & \textbf{83.8} $\pm$ 22.4 &0.07 \\
Home Outdoor (3)  & 72.8 $\pm$ 8.5& 0.31& \textbf{58.1} $\pm$ 12.4 & 0.12\\
Disaster (4) & 252.2 $\pm$ 21.2  & 0.37& \textbf{189.8} $\pm$ 29.6& 0.18\\
 
\hline
\hline
\end{tabular}
\label{table2}
\end{table}

\subsection{Real Robot Experiment}

We implement search for a static target with a heterogeneous three robot team. This preliminary experiment demonstrates that our proposed method is viable in real world environments, however, execution of the search required human intervention due to localization errors. The team is comprised of two Unitree A1 quadrupeds with different sensor suites and a Toyota HSR. One A1 quadruped is equipped with a Velodyne VLP-16 3D Lidar and a RealSense D435. The other is equipped with a RPLidar A3 2D Lidar and a Realsense D435. Both quadrupeds have Intel NUC Mini PCs onboard which communicate with the robot hardware. The HSR is equipped with Hokuyo 2D Lidar, RGB-D camera sensing and has an on board Jetson TK1. The central search server is run on a remote laptop and we use Robofleet \cite{sikand2021robofleet} for efficient communication between the server and agents. 
The entire framework is implemented in ROS Melodic. At initial planning or a re-planning step, the search server generates a set of n paths and then sends those to the agents (see Fig. \ref{fig:api}).

The search experiment is performed in the Aerospace Engineering building at the University of Texas at Austin while looking for a static volleyball and the total mission time is one minute 14 seconds. The experiment is depicted in Fig \ref{fig:asesearch}. 

\section{Concluding Remark}
This paper addresses online search for a general heterogeneous multi agent systems. 
The mission completion time of our method is compared with a state-of-the-art algorithm and shown to outperform it. We further validate our algorithm with extensive simulation results in Gazebo using ROS. Finally, we demonstrate the efficacy of this proposed method in a real robot experiment. Overall results validate the effectiveness and robustness of the proposed algorithm. Several future works remain, however, current efforts are towards including a target motion model for dynamic targets and performing continuous coverage.


\bibliographystyle{IEEEtran}
\bibliography{coverage}

\end{document}